\pdfoutput=1
\documentclass[11pt]{article}

\usepackage{acl}

\usepackage{times}
\usepackage{latexsym}

\usepackage[T1]{fontenc}
\usepackage[utf8]{inputenc}

\usepackage{microtype}

\usepackage{graphicx}
\usepackage{longtable}
\usepackage{todonotes}

\usepackage{xurl}
\usepackage{booktabs}
\usepackage{verbatim}
\usepackage{multirow}
\usepackage{subcaption}

\newcommand{\datasetsnum}{80 } 
\newcommand{\alldatasetsnum}{342 } 
\newcommand{\languagesnum}{27 } 

\title{Assessment of Massively Multilingual Sentiment Classifiers}

\author{
Krzysztof Rajda$^{1,2}$,
Łukasz Augustyniak$^{1,2}$, \\
{\bf Piotr Gramacki$^{2}$},
{\bf Marcin Gruza$^{1,2}$},
{\bf Szymon Woźniak$^{2}$} \\
{\bf Tomasz Kajdanowicz$^{1}$} \\
$^{1}$Department of Artificial Intelligence, \\ Wrocław University of Science and Technology \\
$^{2}$Brand24 AI \\
 \texttt{\{krzysztof.rajda,lukasz.augustyniak\}@pwr.edu.pl} \\
 \texttt{\{piotr.gramacki,marcin.gruza,szymon.wozniak\}@brand24.com} \\
}

\date{}

\begin{document}
\maketitle
\begin{abstract}

Models are increasing in size and complexity in the hunt for SOTA. But what if those 2\% increase in performance does not make a difference in a production use case? Maybe benefits from a smaller, faster model outweigh those slight performance gains. Also, equally good performance across languages in multilingual tasks is more important than SOTA results on a single one. We present the biggest, unified, multilingual collection of sentiment analysis datasets. We use these to assess 11 models and \datasetsnum high-quality sentiment datasets (out of \alldatasetsnum raw datasets collected) in \languagesnum languages and included results on the internally annotated datasets. We deeply evaluate multiple setups, including fine-tuning transformer-based models for measuring performance. We compare results in numerous dimensions addressing the imbalance in both languages coverage and dataset sizes. Finally, we present some best practices for working with such a massive collection of datasets and models from a multilingual perspective. 

\end{abstract}

\section{Introduction}

Multilingual text representations are becoming increasingly important in science as well as the business community. However how universal and versatile they truly are? Can we use them to train one, multilingual, production-ready sentiment classifier? To verify this research question, we gathered a massive collection of sentiment analysis datasets and evaluated 11 different models on them. We want to assess the performance of fine-tuning languages models as well as language models as feature extractors for simpler, even linear models. 

Sentiment analysis is subjective and both domain and language-dependent, hence there is an even greater need to understand the behaviour and performance of the multilingual setup. We focused on multilingual sentiment classification because our business use cases involve the analysis of texts in multiple languages across the world. Moreover, one universal model in a production environment is much easier to deploy, maintain, monitor, remove biases or improve the model's fairness - especially in cases when the load differs between languages and could change over time. We want to compare state-of-the-art multilingual embedding methods and select the ones with the best performance across languages. 

The main objective of this article is to answer the following Research Questions: (RQ1) Are we able to create a single multilingual sentiment classifier, that performs equally well for each language? (RQ2) Does fine-tuning of transformer-based models significantly improve sentiment classification results? (RQ3) What is the relationship between model size and performance? Is bigger always better?

Our main contribution includes 3 main points. Firstly, we perform a large scoping review of published sentiment datasets. Using a set of rigid inclusion and exclusion criteria, we filter the initial pool of \alldatasetsnum datasets down to \datasetsnum high-quality datasets representing \languagesnum languages. Secondly, we evaluated how universal and versatile multilingual text representations are for the sentiment classification problem. Finally, we compared many deep learning-based approaches with fine-tuning and without it for multilingual sentiment classification.

The remainder of this paper is organized as follows: Section 2 presents a literature review on the topic of multilingual sentiment analysis; Section 3 describes the language models, datasets, and our evaluation methodology; Section 4 describes the conducted experiments and summarizes the results; Section 5 discusses the results in terms of research questions; Section 6 presents conclusions and describes further works.

\section{Related Work}

\textbf{Multilingual Text Representations.} 
 
Initially, multilingual text representations were obtained using multilingual word embeddings \cite{ruder2019}. These were created using various training techniques, parallel corpora, and dictionaries, for example by aligning the monolingual Word2Vec \cite{mikolov2013efficient} vector spaces with linear transformations using small parallel dictionaries \cite{mikolov2013exploiting}.

To better represent longer texts, modern approaches use more complex contextual language models like BiLSTM \cite{model:laser} and Transformers \cite{model:labse, model:xlm-r, model:mbert, model:mt5, model:mUSE}. Their multilingual capabilities result from pretraining on multilingual objective tasks like machine translation \cite{model:laser}, translation language modelling (TLM) \cite{model:xlm-r, lample2019crosslingual} or translation ranking \cite{model:labse, yang2019improving}. Details of the models used in our experiments are described in Section \ref{sec:embedding_models}.

The quality of multilingual text representations is usually evaluated with cross- and multilingual tasks like cross-lingual natural language inference \cite{xnli_benchmark}, question answering \cite{lewis2020mlqa}, named entity recognition \cite{conll_benchmark2002, conll_benchmark2003}  or parallel text extraction \cite{zweigenbaum-etal-2017-overview, ziemski-etal-2016-united}. 

Another important benchmark is XTREME \cite{hu2020xtreme}, which is designed for testing the abilities of cross-lingual transfer across 40 languages and 9 tasks. Despite its massive character, XTREME lacks benchmarking task of sentiment analysis, also only mBERT, XLM, XLM-R, and MMTE are used as baseline models. We try to fill this gap with our work.

\citet{related:mbert-cross-lingual-ability} performed extensive research on the cross-lingual ability of mBERT.  \citet{related:mbert-on-low-resource} compared mBERT with monolingual models and found that it under-performs on low-resource languages. \citet{related:mbert-context-vs-non-context} analyzed a cross-lingual ability of mBERT considering a contextual aspect of mBERT and dataset size. There is a significant lack of detailed analysis of characteristics of other language models, despite mBERT.

\textbf{Multilingual Sentiment Analysis.} In literature, there are several examples of reviews, which focus on \textit{traditional} sentiment analysis methods (e.g., lexicon-based, lexical features engineering, shallow models), while not mentioning any embedding-based methods \cite{dashtipour2016multilingual, sagnika2020review}. They are a valuable source of information about sentiment datasets. However, modern NLP applications often utilize deep learning techniques, which were not covered there. An example of a deep learning-based approach was presented by \citet{attia2018}, who trained a convolutional neural network (CNN) on word-level embeddings of texts in English, German and Arabic, a separate model for each language. This approach requires many resources and computations as one has to create a separate embedding dictionary for each language. An alternative approach is to use character-level embeddings. \citet{wehrmann2017} trained such a model to classify tweets written in English, German, Portuguese, and Spanish as either positive or negative. This approach requires fewer parameters than word embedding models.

Newer approaches to multilingual sentiment analysis use deep models and machine translation e.g. \citet{can2018} trained a Recurrent Neural Network (RNN) on English reviews and evaluated it on machine-translated reviews in Russian, Spanish, Turkish and Dutch. They used the Google Translation API and pre-trained GloVe embeddings for English. \citet{KANCLERZ2020128} used LASER sentence embeddings to learn a sentiment classifier on Polish reviews and used this classifier to predict sentiment on reviews translated into other languages. As we can see most of the research covers only a couple of languages for sentiment analysis. Hence, we decided to gather a massive collection of \alldatasetsnum datasets in \languagesnum languages.

\begin{table*}[t]
\centering
\caption{Models used in experiments - inference times, number of parameters, and languages used in pre-training, base model and data used in pre-training}
\label{tab:models-technical}
\resizebox{\textwidth}{!}{%
\begin{tabular}{lrrrlll}
\toprule
Model & Inf. time [s] & \#params & \#langs & base$^a$ & data & reference \\%& data size$^f$ \\
\midrule
mT5 &  1.69  & 277M & 101 & T5 & CC$^b$ & \cite{model:mt5} \\%&  6.6B p. \\
LASER &  1.64  &  52M & 93 & BiLSTM & OPUS$^c$ & \cite{model:laser} \\%& 223M sp. \\
mBERT &  1.49 &  177M & 104 & BERT & Wiki & \cite{model:mbert} \\%& ??? \\
MPNet** &  1.38  & 278M  & 53 & XLM-R & OPUS$^c$, MUSE$^d$, Wikititles$^e$ & \cite{model:multi-sbert} \\%& ??? \\
XLM-R-dist** &  1.37  & 278M & 53 & XLM-R & OPUS$^c$, MUSE$^d$, Wikititles$^e$ & \cite{model:multi-sbert} \\%& ??? \\
XLM-R &  1.37 & 278M & 100 & XLM-R & CC & \cite{model:xlm-r} \\%& 2,5TB \\
LaBSE &  1.36 & 470M & 109 & BERT & CC, Wiki + mined bitexts & \cite{model:labse} \\%& 17B s. + 6B sp.\\
DistilmBERT &  0.79  &  134M & 104 & BERT & Wiki & \cite{model:distilmbert} \\%& ??? \\
mUSE-dist** &  0.79  &  134M & 53 & DistilmBERT & OPUS$^c$, MUSE$^d$, Wikititles$^e$ & \cite{model:multi-sbert} \\%& ??? \\
mUSE-transformer* &  0.65  &  85M & 16 & transformer & mined QA + bitexts, SNLI & \cite{model:mUSE} \\%& 60M+ sp. per lang\\
mUSE-cnn* &  0.12 &  68M & 16 & CNN & mined QA + bitexts, SNLI & \cite{model:mUSE} \\%& 60M+ sp. per lang\\
\bottomrule
\end{tabular}
}
\caption*{\footnotesize{*mUSE models were used in TensorFlow implementation in contrast to others in torch $^a$ Base model is either monolingual version on which it was based or another multilingual model which was used and adopted $^b$ Colossal Clean Crawled Corpus in multilingual version (mC4) $^c$ multiple datasets from OPUS website (https://opus.nlpl.eu), $^d$ bilingual dictionaries from MUSE (https://github.com/facebookresearch/MUSE), $^e$ just titles from wiki articles in multiple languages}}
\end{table*}

\section{Evaluation Methodology}

We conducted several experiments to answer if there is a truly universal multilingual text representation model (Table \ref{tab:models-technical}). We tested their performance based on the largest sentiment analysis dataset in the literature.

\subsection{Multilingual Language Models}
\label{sec:embedding_models}

We used multiple language models as text representation methods (Table \ref{tab:models-technical}). We aimed to select models varied in terms of architecture, size, and type of data used in pre-training. We selected two models which do not use transformer architecture (CNN and BiLSTM) as a baseline. We also used models, based on multiple different transformer architectures (T5, BERT, RoBERTa). We also included models' trained with multilingual knowledge distillation \cite{model:multi-sbert} such as \textit{paraphrase-xlm-r-multilingual-v1} (XLM-R-dist), \textit{distiluse-base-multilingual-cased-v2} (mUSE-dist), \textit{paraphrase-multilingual-mpnet-base-v2} (MPNet). We also included models trained on multilingual corpus like Wikipedia (Wiki) or Common Crawl (CC) as well as models trained with the use of parallel datasets. Selected models differ in size - from LASER with 52M parameters to LaBSE with 470M. They also differ regarding covered languages, from 16 up to more than a hundred. By a number of languages, we mean how many were used to create a specific model, not all languages supported by the model (an example is MPNet, trained using 53 languages, but as it is based on XLM-R, it supports 100). We also compared inference time which was calculated as a mean of inference times of 500 randomly selected texts samples from all datasets. The hardware used is described in Section \ref{sec:hardware}. We searched for models comparison in similar tasks in literature but failed to find any, which compares more than 2 or 3 models. All models used are characterized in Table \ref{tab:models-technical}.

\begin{table*}[t]
\centering
\caption{Summary of \datasetsnum high-quality datasets selected. Categories: N - News, O - Other, R - Reviews, SM - Social Media}
\label{tab:dataset_languages_3_class}
\begin{tabular}{lc|cccc|rrr|cc}
\toprule
{} & Count & \multicolumn{4}{c}{Category} & \multicolumn{3}{c}{Samples} & \multicolumn{2}{c}{Mean \#} \\
{} &   &  N & O & R & SM &      NEG &     NEU &      POS &       words &   characters \\
%language &       &          &       &         &              &          &         &          &             &              \\
\midrule
English    &    17 &        3 &     4 &       4 &            6 &  305,782 &  289,847 &  1,734,857 &    42 &        233 \\
Arabic     &     9 &        0 &     1 &       4 &            4 &  139,173 &  192,463 &    600,439 &    28 &        159 \\
Spanish    &     5 &        0 &     1 &       3 &            2 &  110,156 &  120,668 &    188,068 &   145 &        864 \\
Chinese    &     2 &        0 &     0 &       2 &            0 &  118,023 &   68,953 &    144,726 &    48 &          - \\
German     &     6 &        0 &     0 &       1 &            5 &  105,416 &   99,291 &    111,180 &    19 &        131 \\
Polish     &     4 &        0 &     0 &       2 &            2 &   78,309 &   61,041 &     97,338 &    39 &        245 \\
French     &     3 &        0 &     0 &       1 &            2 &   84,324 &   43,097 &     83,210 &    19 &        108 \\
Japanese   &     1 &        0 &     0 &       1 &            0 &   83,985 &   41,976 &     83,819 &    60 &          - \\
Czech      &     4 &        0 &     0 &       2 &            2 &   39,687 &   59,181 &     97,419 &    29 &        168 \\
Portuguese &     4 &        0 &     0 &       0 &            4 &   57,737 &   54,145 &     45,952 &    12 &         73 \\
Slovenian  &     2 &        1 &     0 &       0 &            1 &   34,178 &   50,055 &     29,310 &   161 &       1054 \\
Russian    &     2 &        0 &     0 &       0 &            2 &   32,018 &   47,852 &     31,060 &    11 &         73 \\
Croatian   &     2 &        1 &     0 &       0 &            1 &   19,907 &   19,298 &     38,389 &    86 &        556 \\
Serbian    &     3 &        0 &     0 &       2 &            1 &   25,580 &   31,762 &     19,026 &   176 &       1094 \\
Thai       &     2 &        0 &     0 &       1 &            1 &    9,327 &   28,615 &     34,377 &    18 &        317 \\
Bulgarian  &     1 &        0 &     0 &       0 &            1 &   14,040 &   28,543 &     19,567 &    12 &         85 \\
Hungarian  &     1 &        0 &     0 &       0 &            1 &    9,004 &   17,590 &     30,088 &    11 &         83 \\
Slovak     &     1 &        0 &     0 &       0 &            1 &   14,518 &   12,735 &     29,370 &    13 &         97 \\
Albanian   &     1 &        0 &     0 &       0 &            1 &    6,958 &   14,675 &     22,651 &    13 &         90 \\
Swedish    &     1 &        0 &     0 &       0 &            1 &   16,664 &   12,912 &     11,770 &    14 &         94 \\
Bosnian    &     1 &        0 &     0 &       0 &            1 &   12,078 &   11,039 &     13,066 &    12 &         75 \\
Urdu       &     1 &        0 &     1 &       0 &            0 &    5,244 &    8,580 &      5,836 &    13 &         69 \\
Hindi      &     1 &        0 &     0 &       0 &            1 &    4,992 &    6,392 &      5,615 &    27 &        128 \\
Persian    &     1 &        0 &     0 &       1 &            0 &    1,619 &    5,074 &      6,832 &    21 &        104 \\
Italian    &     2 &        0 &     0 &       0 &            2 &    4,043 &    4,193 &      3,829 &    16 &        104 \\
Hebrew     &     1 &        0 &     0 &       0 &            1 &    2,283 &      238 &      6,098 &    22 &        110 \\
Latvian    &     1 &        0 &     0 &       0 &            1 &    1,379 &    2,617 &      1,794 &    20 &        138 \\
\bottomrule
\end{tabular}
\end{table*}

\subsection{Datasets}
\label{sec:benchmarks-datasets}

We gathered \alldatasetsnum sentiment analysis datasets containing texts from multiple languages, data sources and domains to check our research questions. We searched for datasets in various sources, like Google Scholar, GitHub repositories, and the HuggingFace datasets library. Such a large number of datasets allows us to estimate the quality of language models in various conditions with greater certainty. To the best of our knowledge, this is the largest sentiment analysis datasets collection currently gathered and researched in literature. After preliminary analysis, we selected \datasetsnum datasets of reasonable quality based on 5 criteria. (1) We rejected datasets containing weak annotations (e.g., datasets with labels based on emoji occurrence or generated automatically through classification by machine learning models), as our analysis showed that they may contain too much noise \cite{noise-ratio}. (2) We reject datasets without sufficient information about the annotation procedure (e.g., whether annotation was manual or automatic, number of annotators) because it is always a questionable decision to merge datasets created with different annotation guidelines. (3) We accepted reviews datasets and mapped their rating labels to sentiment values. The mapping rules are described in section \ref{sec:processing}. (4) We rejected 2-class only datasets (positive/negative without neutral), as our analysis showed their low quality in terms of 3-class usage. (5) Some datasets contain samples in multiple languages - we split them and treated each language as a separate dataset.

\subsubsection{Data Preprocessing}
\label{sec:processing}

Working with many datasets means that they could contain different types of text, various artefacts such as URL or HTML tags, or just different sentiment classes mappings. We applied a couple of preprocessing steps to each dataset to unify all datasets. We dropped duplicated texts. We removed URLs, Twitter mentions, HTML tags, and emails. During an exploratory analysis, we spotted that review-based datasets often contain many repeated texts with contradictory sentiment scores. We deduplicated such cases and applied a majority voting to choose a sentiment label. Finally, we unified labels from all datasets into 3-class (negative, neutral, positive). In the case of datasets containing user ratings (on a scale of 1-5) along with their review texts, we mapped the ratings to sentiment as follows: the middle value (3) of the rating scale was treated as a neutral sentiment, ratings below the middle as negative sentiment, and ratings above the middle as positive sentiment.

Presenting statistics of \datasetsnum datasets across \languagesnum languages could be challenging. We checked different aggregating and sorting of datasets to make their statistics as readable as possible and easily usable for results discussion. We decided to group datasets by their language and next sorted them based on the number of examples in every aggregate - Table \ref{tab:dataset_languages_3_class}. 
In total, we selected \datasetsnum datasets containing 6,164,942 text samples. Most of the texts in the datasets are in English (2,330,486 samples across 17 datasets), Arabic (932,075 samples across 9 datasets), and Spanish (418,892 samples across 5 datasets). The datasets contain text from various categories: social media (44 datasets), reviews (24 datasets),  news (5 datasets), and others (7 datasets). They also differ in the mean number of words and characters in examples. See the detailed information of datasets used is in Tables \ref{tab:dataset_monolingual} and \ref{tab:dataset_multilingual}.

We also selected around 60k samples for training and validation and another 60k for testing. This is enough for training a small classifier on top of a frozen embedding or fine-tuning a transformer-based model (see Section \ref{sec:experimental_setup}). This was also done due to computation resources limitations.

\begin{table}
    \centering
    \caption{Statistics of the internal dataset}
    \label{tab:internal-dataset}
    \begin{tabular}{l|rrrr}
    \toprule
    lang & samples & NEG  & NEU  & POS  \\ \midrule
    pl   & 2968    & 14\% & 60\% & 26\% \\
    en   & 943     & 4\% & 74\% & 22\% \\ \bottomrule
    \end{tabular}
\end{table}

\subsubsection{Internal Dataset}
\label{sec:internal-dataset}

We have also used an internal dataset that was manually annotated. It is multi-domain and consists of texts from various Internet sources in Polish and English. It includes texts from social media, news sites, blogs and forums. We used this dataset as a gold standard. We need it because we do not know exact annotation guidelines from literature datasets and we assume that those guidelines differed between datasets. In our gold dataset, each text was annotated by 3 annotators with majority label selection. The annotators achieved a high agreement measured by Cohen's kappa: $0.665$ and Krippendorff's alpha: $0.666$. Statistics of this dataset are presented in Table \ref{tab:internal-dataset}. All samples were trimmed to the length of 350 chars (mean length of 145 chars).

\subsection{Experimental Scenarios}
\label{sec:experimental_setup}

We wanted to compare multilingual models in different use cases. Firstly, we wanted to see how much information is stored in pre-trained embedding. In this scenario, we used each of the text representations models listed in Section \ref{sec:embedding_models} as a feature extractor and coupled them with only a small linear classification head. We used an average from a final layer as a text representation.  We will refer to this scenario \textbf{Just Head - Linear}. In the second scenario, we replaced a linear classifier with a BiLSTM classifier, still using the text representation model as a feature extractor. We fed BiLSTM layer with outputs from the last layer of the feature extractor (\textbf{Just Head - BiLSTM}). LASER and mUSE do not provide per-token embeddings and therefore, were not included in this scenario. Since most of our models are transformer-based, we decided to test them in a fine-tuning setup. This last scenario evaluated the fine-tuning of all transformer-based models (referred to as \textbf{fine-tuning}), with an exception made for mUSE-transformer because it was not possible to do with our implementation in PyTorch with Huggingface models.

For each scenario, we prepared 3 test metrics, which we refer to as a \textit{whole test}, \textit{average by dataset} and \textit{internal}. Each of them separately measures model performance but all of them are based on a macro F1-score. The \textit{whole test} is calculated on all samples from datasets described in \ref{sec:benchmarks-datasets} combined. It is meant to reflect the real-life performance of a model because our real-world applications often deal with an imbalance in languages distribution (with English being the most popular language used on the Internet). On \textit{average by dataset}, we first calculate the macro F1-score on each dataset and then calculate the average of those scores. This is meant to show whether the model was not too over-fitted for the majority of languages or the biggest datasets. Finally, in the \textit{internal} scenario, we assess them on our internal dataset (described in \ref{sec:internal-dataset}) to measure performance in our domain-specific examples.

\subsection{Evaluation Procedure}

To show how each model performs in a bird's eye view, we prepared Nemenyi diagrams~\cite{nemenyi1963distribution} for all three experimental setups. Nemeneyi post-hoc statistical test finds groups of models that differ. It was used on the top of multiple comparisons Friedman test \cite{Demsar2006}. The Nemeneyi test makes a pair-wise comparison of all model's ranks. We used alpha equal to 5\%. The Nemeneyi test provides critical distance for compared groups that are not significantly different from each other. 

\subsection{Models Setup}

For each scenario, we adjusted hyperparameters. The hidden size was set to LM's embedding size for linear and fine-tuning and 32 for BiLSTM. By hidden size, we mean middle linear layer size, or in the case of BiLSTM - its hidden size parameter. BiLSTM uses a smaller hidden size because our experiments showed that it does not hurt performance but increases efficiency. The learning rate was initially the same for all scenarios, at the well-established value of 1e-3. We then modified it for each version by decreasing it for fine-tuning (to 1e-5) and slightly increasing it for BiLSTM based model (5e-3). The batch size was determined by our GPU's memory size. We used 200 for linear and BiLSTM and 6 for fine-tuning. We used dropout in classification head - 0.5 for BiLSTM and 0.2 for other scenarios. We trained our models for 5 epochs in the fine-tuning scenario and 15 in two others, as those were the max number of epochs before the models started overfitting. We tested with the best F1-score on a validation dataset.

\section{Results}

We divided our results into three layers. Firstly, we show a general bird's eye view of all models - it helps to spot the best and the worst models. Then, we provide detailed results for each model aggregated per dataset. Finally, to dig deeper into the model's performance, we show numerical results for each model for each language.

\begin{figure}[t]
    \centering
    \begin{subfigure}{0.48\textwidth}
        \includegraphics[width=\textwidth]{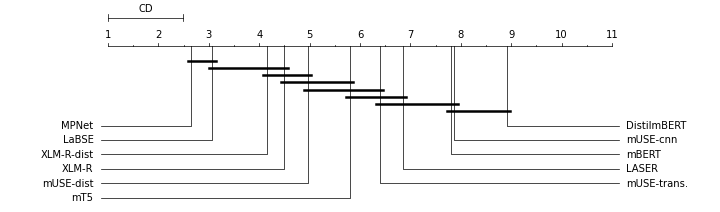}
        \caption{Just head - Linear}
        \label{fig:nemenyi-linear}
    \end{subfigure}
    \begin{subfigure}{0.48\textwidth}
        \includegraphics[width=\textwidth]{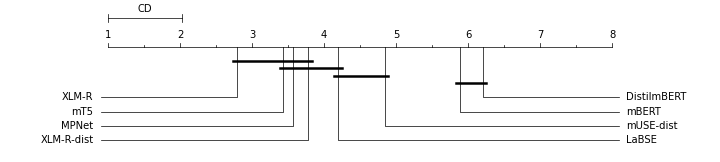}
        \caption{Just head - BiLSTM}
        \label{fig:nemenyi-bilstm}
    \end{subfigure}
    \begin{subfigure}{0.48\textwidth}
        \includegraphics[width=\textwidth]{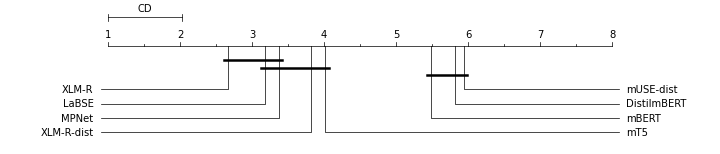}
        \caption{Fine-tuning}
        \label{fig:nemenyi-finetuning}
    \end{subfigure}
    \caption{Nemenyi diagrams based on the ranking of models according to the F1-score on each dataset}
    \label{fig:nemenyi-full}
\end{figure}

\begin{table}[t]
	\centering
	\caption{Aggregated results of models (F1 score in \%). The best results for each test set are highlighted. (W - whole test, A - avg. by dataset, I - internal)}
	\label{tab:finetuning-agg-results}
	\resizebox{\columnwidth}{!}{%
		\begin{tabular}{c|rrrrrrrrrrr}
			\toprule
			  & \rotatebox{90}{XLM-R} & \rotatebox{90}{LaBSE} & \rotatebox{90}{MPNet} & \rotatebox{90}{XLM-R-dist} & \rotatebox{90}{mT5} & \rotatebox{90}{mBERT} & \rotatebox{90}{DistilmBERT} & \rotatebox{90}{mUSE-dist} & \rotatebox{90}{LASER} & \rotatebox{90}{mUSE-trans.} & \rotatebox{90}{mUSE-cnn} \\
			\midrule & \multicolumn{11}{c}{Just Head - Linear} \\
			\midrule
			W & 62                    & 62                    & \textbf{63}           & 60                         & 59                  & 56                    & 55                          & 59                        & 55                    & 55 & 54          \\
			A & 51                    & 54                    & \textbf{55}           & 51                         & 49                  & 45                    & 43                          & 50                        & 47                    & 47 & 45          \\
			I & 55                    & \textbf{61}           & \textbf{61}           & 56                         & 50                  & 43                    & 38                          & 60                        & 50                    & 49 & 50          \\
			\midrule & \multicolumn{11}{c}{Just Head - BiLSTM} \\
			\midrule
			W & \textbf{66}           & 62                    & 63                    & 62                         & 65                  & 60                    & 59                          & 62                        & -                     & -  & -           \\
			A & \textbf{57}           & 55                    & 56                    & 54                         & 56                  & 49                    & 48                          & 54                        & -                     & -  & -           \\
			I & \textbf{64}           & 63                    & \textbf{64}           & 63                         & 63                  & 54                    & 48                          & \textbf{64}               & -                     & -  & -           \\
			\midrule & \multicolumn{11}{c}{Fine-tuning} \\
			\midrule
			W & \textbf{68}           & \textbf{68}           & 67                    & 67                         & 66                  & 65                    & \textbf{64}                 & 63                        & -                     & -  & -           \\
			A & 61                    & \textbf{62}           & \textbf{62}           & \textbf{62}                & 60                  & 56                    & 56                          & 56                        & -                     & -  & -           \\
			I & \textbf{70}           & 69                    & 65                    & 67                         & 67                  & 57                    & 58                          & 60                        & -                     & -  & -           \\
			\midrule
		\end{tabular}
	}
\end{table}

\subsection{Bird’s Eye View}

There is no significantly best embedding model in any of the tested scenarios based on the Nemenyi diagrams - Figure \ref{fig:nemenyi-full}. However, we can see that the MPNet proved to be the best (for the linear scenario) and not significantly worse than the best - XLR-M model - in the other two scenarios. It is also worth mentioning that mBERT-based models (mBERT and DistillmBERT) proved to be the worst language models for our tasks.

\subsection{Aggregated by Dataset}

All models achieve better results with fine-tuning (up to 0.7 F1-score) than with extraction of vectors from text and then applying linear (up to 0.61) or BiLSTM (up to 0.64) layers, shows Table \ref{tab:finetuning-agg-results}. The performance gains are higher when fine-tuning models pretrained on MLM and TLM tasks (like mBERT or XLM-R) compared to models, which were trained with sentence classification tasks, sentence similarity tasks or similar (like LaBSE). For example, mBERT had gains of 9, 11, and 14 percentage points (pp) on \textit{whole test}, \textit{average by dataset} and \textit{internal} test cases, DistilmBERT - 9, 13 and 20pp, XLM-R - 6, 10, and 15pp. At the same time, LaBSE had only 6, 8, and 7pp and MPNet - 4, 7, 4pp. Still, those models achieve better overall performance. Fine-tuning reduces inequalities in the results between models (0.55 vs 0.43 for best and worst models in Just head - Linear setup, and 0.62 vs 0.56 after Fine-tuning for \textit{average by dataset} metric). Those results were meant to show a general comparison between fine-tuned models against training just classification head. 

The additional BiLSTM layer on top of transformer token embeddings improves the results of the model with only a linear layer in most cases. The differences are most clear in the case of the results for our internal dataset, where the result improved even by 13pp. (from 50\% to 63\%) for the mT5 model.

Those results show, that three models are the most promising choices: XLM-R, LaBSE and MPNet. They achieve comparable performance in all scenarios and test cases. Furthermore, they are better than other models in almost all test cases. XLM-R-dist was very close to those, but analysis with Nemenyi diagrams shows that it is slightly worse than those three. 

\begin{figure*}
    \centering
    \includegraphics[width=\textwidth]{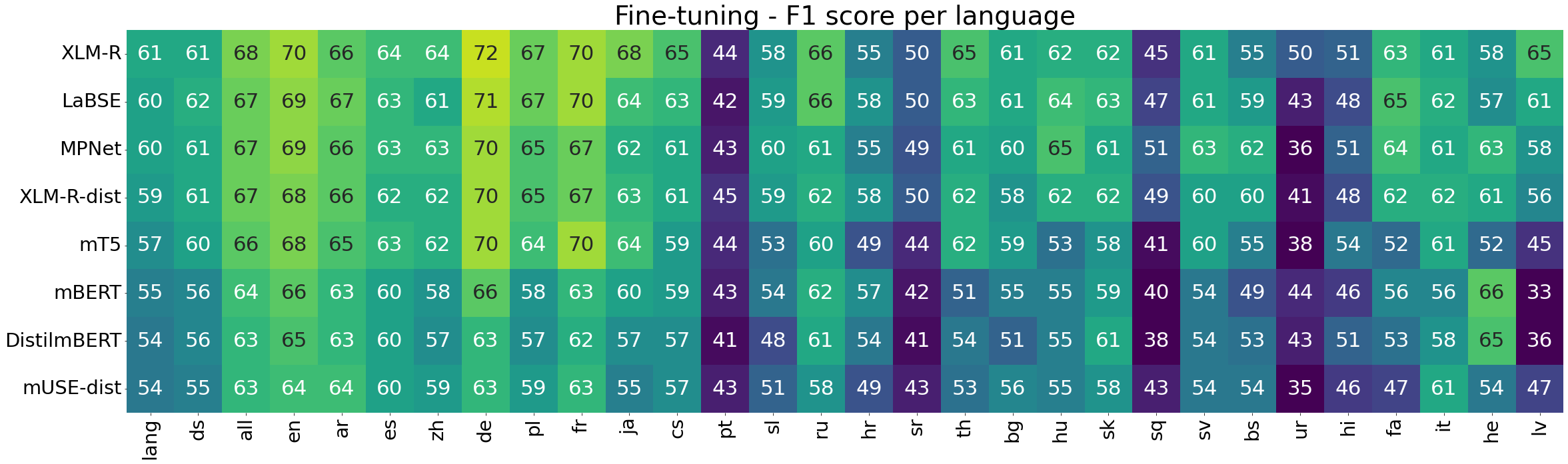}
    \caption{Detailed results of models' comparison.}
    \caption*{\footnotesize{Legend:
    \textbf{lang} - averaged by all languages,
    \textbf{ds} - averaged by dataset,  \textbf{ar} - Arabic, \textbf{bg} - Bulgarian, \textbf{bs} - Bosnian, \textbf{cs} - Czech, \textbf{de} - German, \textbf{en} - English, \textbf{es} - Spanish, \textbf{fa} - Persian, \textbf{fr} - French, \textbf{he} - Hebrew, \textbf{hi} - Hindi, \textbf{hr} - Croatian, \textbf{hu} - Hungarian, \textbf{it} - Italian, \textbf{ja} - Japanese, \textbf{lv} - Latvian, \textbf{pl} - Polish, \textbf{pt} - Portuguese, \textbf{ru} - Russian, \textbf{sk} - Slovak, \textbf{sl} - Slovenian, \textbf{sq} - Albanian, \textbf{sr} - Serbian, \textbf{sv} - Swedish, \textbf{th} - Thai, \textbf{ur} - Urdu, \textbf{zh} - Chinese.}}
    \label{fig:finetuning-languages-f1}
\end{figure*}

\subsection{Every Model for Every Language}

We assessed the performance of each model in each experimental scenario concerning the language. The texts were sub-sampled with stratification by language and class label so that language distribution in the test dataset reflects this in the whole dataset. It means that some languages are underrepresented. We also include the total Macro F1 score value in column "all". Results are presented in Figure \ref{fig:finetuning-languages-f1} for fine-tuning scenario and in Figure \ref{fig:finetuning-vs-static-results} for others. Those results confirm conclusions from the previous section about the advantage of XLM-R, LaBSE and MPNet. They have the best performance in most languages and together with XLM-R-dist, there are no big differences between them.

\section{Discussion}

\paragraph{RQ1: Are we able to create a single multilingual sentiment classifier, performing equally well for each language?}

When considering only the best models (XLM-R, LaBSE, MPNet) in the fine-tuning setup, we observed that they achieve best or close to best results in every language - Figure \ref{fig:finetuning-vs-static-results}. In some languages, results are significantly worse than in others, but this is also true for other models evaluated as it may be caused by differences in the number of samples, quality, and difficulty of samples in those languages. Therefore, we can say that one model can work exceptionally well in all languages. On the other hand, statistical analysis which is presented in the form of Nemenyi diagrams in Figures \ref{fig:nemenyi-linear}, \ref{fig:nemenyi-bilstm} and \ref{fig:nemenyi-finetuning} showed that there is no statistical difference between top models in fine-tuning setup, so it is not possible to state which of those is the best one. We can rather state which group of models proved to be significantly better than others. 

\begin{figure}
    \centering
    \includegraphics[width=\columnwidth]{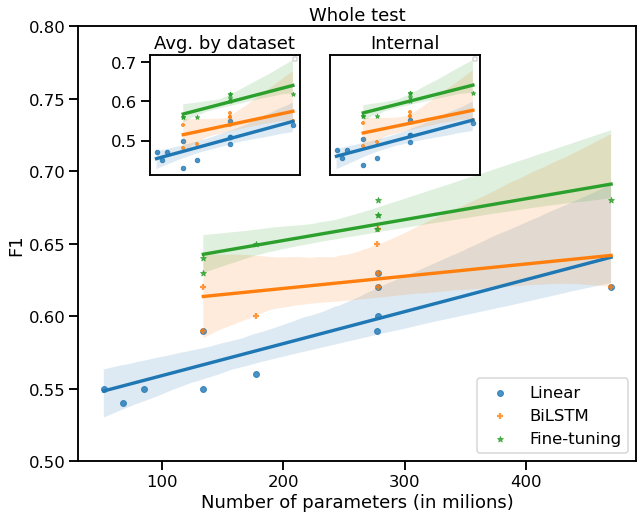}
    \caption{Results for models by their size and scenario.}
    \label{fig:f1_by_model_size}
\end{figure}

\paragraph{RQ2: Does fine-tuning of transformer-based models significantly improve sentiment classification results?}

All models worked better when fine-tuned, but the performance gain varied from one to another. It was between 4 (mUSE-dist) and 9 (mBERT and DistilmBERT) pp. F1 on the benchmark test dataset, and between 0 (mUSE-dist) and 20 pp. (DistilmBERT) on our internal dataset. The 17, 15, and 14 pp. gain of mT5, XLM-R, and DistilmBERT on the internal dataset is also worth noting. In general, the most significant gain can be observed in models trained on language modelling only (MLM or TLM), such as XLM-R and mBERT.

\paragraph{RQ3: What is the relationship between model size and performance? Is bigger always better?}

The results of our experiments showed that there exists a correlation between the classification result of the language model with its number of parameters. Figure \ref{fig:f1_by_model_size} shows that, for all scenarios and test dataset types, bigger models achieve better performance in most cases. However, there are some counterexamples, e.g., mUSE-dist is smaller than mBERT but achieves better performance in Just head - Linear setup, for all dataset types. This indicates that the size of the model is an important factor in its performance, but other factors, like the domain and the type of pretraining task, may also affect the results. Moreover, we observed that this correlation is weaker after fine-tuning.
We can often find the model with similar performance to the best one but significantly smaller and faster for the production environment. 

\paragraph{Your Dataset Splits Matter}

\begin{figure}[t]
    \centering
    \includegraphics[width=\columnwidth]{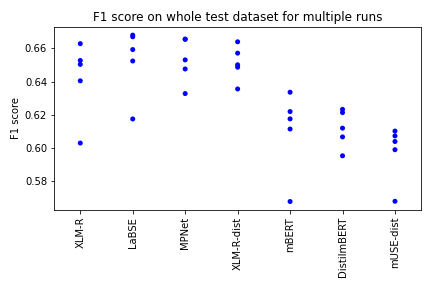}
    \caption{Results of multiple runs of fine-tuning experiments with different seeds.}
    \label{fig:finetuning-seeds-details}
\end{figure}

To determine which model works best, we repeated fine-tuning five times to remove a right/wrong random seed factor for each model and dataset subsampling. Due to computation resources limitations, we selected eight models available in Huggingface for fine-tuning. Interestingly, we can see that one of the samples looks like the outlier - Figure \ref{fig:finetuning-seeds-details} for almost all the evaluated models. The F1-score for this sample is even 4 percentage points worse than other samples' scores. We investigated this anomaly and spotted that it is always the same sample (the same seed for sample generation). As a reminder, since we collected a massive dataset and had limited computational resources, we sub-sampled texts for each of the five runs. Sub-samples between different models stay the same. It looks like the mentioned sample was more difficult than others or had distinctive characteristics. It is hard to tell why without in-depth analysis, hence we intend to conduct further research on the topic of data quality in sentiment analysis tasks using techniques like noise ratio \cite{noise-ratio} or data cartography \cite{dataset-cartoghraphy}. Here, we see an outstanding example of how vital the dataset's preparation could be regarding split for train/dev/test sets. 

\section{Conclusions and Further Works}

In this work we evaluated multilingual text representations for the task of sentiment classification by comparing multiple approaches, using different deep learning methods. In the process, we gathered the biggest collection of multi-lingual sentiment datasets - 80 datasets for 27 languages. We evaluated 11 models (language models and text vectorization techniques) in 3 different scenarios. We found out that it is possible to create one model which achieves the best or most competitive results in all languages in our collected dataset, but there is no statistical difference between top-performing models. We found out that there is a significant benefit from fine-tuning transformer-based language models and that a model size is correlated with performance.

While conducting experiments we identified further issues which we find worth addressing. Dataset quality assessment is in our opinion the most important one and we are planning to address it in further works. Meanwhile, we used datasets with a literature background and trust that they were carefully prepared and have decent quality annotations. We also found out that it is difficult to propose a coherent experiments methodology with such imbalance in languages and datasets sizes. Moreover, analyzing results is difficult, when one must address dimensions of datasets, languages, data sources, models, and experiments scenarios.

Finally, we found out that when sub-sampling a dataset for experiments, seeds play a significant role (see results in Figure \ref{fig:finetuning-seeds-details}). To analyze this phenomenon, we intend to launch further research and use noise ratio \cite{noise-ratio} and data cartography \cite{dataset-cartoghraphy} to understand how this split differs from the others. This will be, in our opinion, a good start to a comprehensive analysis of datasets quality for the multi-lingual sentiment classification task which we intend to perform.

\section*{Acknowledgements}
The work was partially supported by the Department of Artificial Intelligence at Wroclaw University of Science and Technology, and by European Regional Development Fund (ERDF) in RPO WD 2014-2020 (project no. RPDS.01.02.02-02-0065/20). We want to thank Mikołaj Morzy for an initial review and feedback. We want to thank our annotators team - Barbara Orłowska, Daria Szałamacha, Konrad Gajewski and Paweł Odrowąż-Sypniewski.   

\bibliography{acl2020,related_works}
\bibliographystyle{acl_natbib}

% \newpage
\appendix

\section{Appendices}
\label{sec:appendix}

\subsection{Hardware and Software}
\label{sec:hardware}

We performed our experiments using Python 3.9 and PyTorch (1.8.1) (and Tensorflow (2.3.0) for original mUSE). Our experimental setup consists of Intel(R) Xeon(R) CPU E5-2630 v4 @ 2.20GHz and Nvidia Tesla V100 16GB.

\subsection{Detailed Datasets Information}

We present detailed lists of datasets included in our research in Tables \ref{tab:dataset_monolingual} and \ref{tab:dataset_multilingual}. They include language, category, dataset size, class balance and basic texts characteristics.

\subsection{Full Results for Languages}

\begin{figure*}
    \centering
    \includegraphics[width=\textwidth]{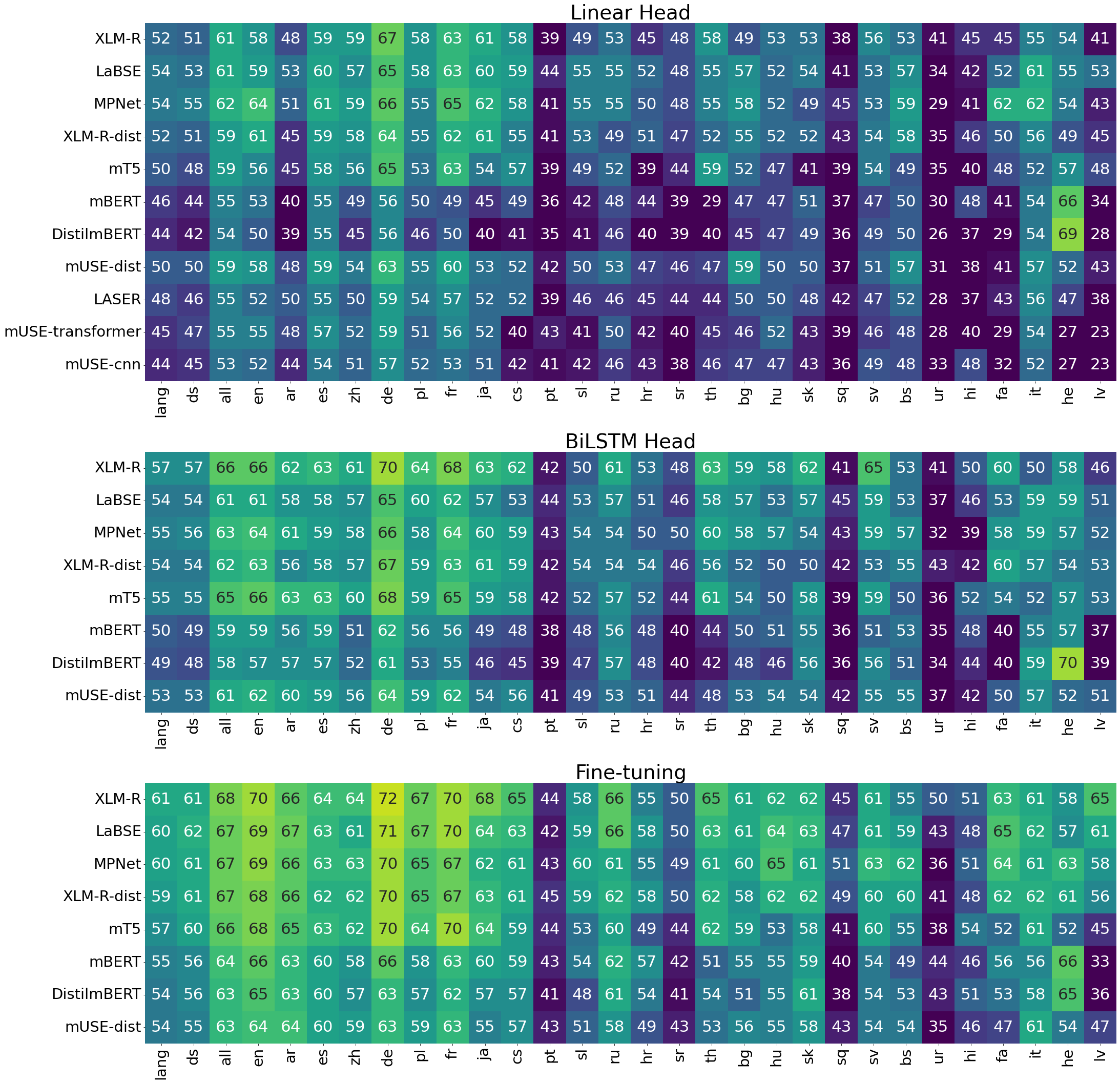}
    \caption{Detailed results of models' comparison. Legend:
    lang - averaged by all languages,
    ds - averaged by dataset, ar - Arabic, bg - Bulgarian, bs - Bosnian, cs - Czech, de - German, en - English, es - Spanish, fa - Persian, fr - French, he - Hebrew, hi - Hindi, hr - Croatian, hu - Hungarian, it - Italian, ja - Japanese, lv - Latvian, pl - Polish, pt - Portuguese, ru - Russian, sk - Slovak, sl - Slovenian, sq - Albanian, sr - Serbian, sv - Swedish, th - Thai, ur - Urdu, zh - Chinese.}
    \label{fig:finetuning-vs-static-results}
\end{figure*}

We include full results of our experiments with results for each language in Figure \ref{fig:finetuning-vs-static-results}. Part with finetuning results was presented earlier in Figure \ref{fig:finetuning-languages-f1}.

% \clearpage
% \onecolumn
% \setlength\LTleft{0pt}
% \setlength\LTright{0pt}

\begin{table*}
\caption{List of all monolingual datasets used in experiments. Category (Cat.): R - Reviews, SM - Social Media, C - Chats, N - News, P - Poems, M - Mixed. HL - human labeled, \#Words and \#Chars are mean values}
\label{tab:dataset_monolingual}
\resizebox{\textwidth}{!}{%
\begin{tabular}{c|c|c|c|r|r|r|r}

 Paper                     &   Lang &   Cat.  & HL &  Samples & NEG/NEU/POS &  \#Words &  \#Char. \\
 \hline
\cite{dataset_ar_oclar} & ar & R &            No &     3096 &        13.0/10.2/76.8 &            9 &           51 \\
\cite{dataset_ar_hard} & ar & R &            No &   400101 &        13.0/19.9/67.1 &           22 &          127 \\
\cite{dataset_ar_labr} & ar & R &            No &     6250 &        11.6/17.9/70.5 &           65 &          343 \\
\cite{dataset_ar_brad} & ar & R &            No &   504007 &        15.4/21.0/63.6 &           77 &          424 \\
\cite{dataset_arsentdl} & ar & SM &           Yes &     2809 &        47.2/23.9/29.0 &           22 &          130 \\
\cite{dataset_ar_astd} & ar & SM &           Yes &     3224 &        50.9/25.0/24.1 &           16 &           94 \\
\cite{dataset_ar_bbn} & ar & SM &           Yes &     1199 &        48.0/10.5/41.5 &           11 &           51 \\
\cite{dataset_ar_bbn} & ar & SM &           Yes &     1998 &        67.5/10.1/22.4 &           20 &          107 \\
\cite{dataset_cz_social_media} & cs & R &            No &    91140 &        32.4/33.7/33.9 &           50 &          311 \\
\cite{dataset_cz_social_media} & cs & R &            No &    92758 &         7.9/23.4/68.7 &           20 &          131 \\
\cite{dataset_cz_social_media} & cs & SM &           Yes &     9752 &        20.4/53.1/26.5 &           10 &           59 \\
\cite{dataset_cz_social_media} & cs & SM &           Yes &     2637 &         30.8/60.6/8.6 &           33 &          170 \\
\cite{dataset_de_sb10k} & de & SM &           Yes &     9948 &        16.3/59.2/24.6 &           11 &           86 \\
\cite{dataset_de_omp} & de & SM &           Yes &     3598 &         47.3/51.5/1.2 &           33 &          237 \\
\cite{dataset_en_silicone} & en & C &           Yes &    12138 &        31.8/46.5/21.7 &           12 &           48 \\
\cite{dataset_en_silicone} & en & C &           Yes &     4643 &        22.3/48.9/28.8 &           15 &           71 \\
\cite{dataset_en_financial_phrasebank_sentences_75agree} & en & N &           Yes &     3448 &        12.2/62.1/25.7 &           22 &          124 \\
\cite{dataset_en_per_sent} & en & N &           Yes &     5333 &        11.6/37.3/51.0 &          388 &         2129 \\
\cite{dataset_en_vader} & en & N &            No &     5190 &        29.3/52.9/17.8 &           17 &          104 \\
\cite{dataset_en_poem_sentiment} & en & P &           Yes &     1052 &        18.3/15.8/65.9 &            7 &           37 \\
\cite{dataset_en_vader} & en & R &            No &     3708 &        34.2/19.5/46.3 &           16 &           87 \\
\cite{dataset_en_vader} & en & R &            No &    10605 &         49.6/1.5/48.9 &           19 &          111 \\
\cite{dataset_en_amazon} & en & R &            No &  1883238 &          8.3/8.0/83.7 &           70 &          382 \\
\cite{dataset_sanders} & en & SM &           Yes &     3424 &        16.7/68.1/15.2 &           14 &           97 \\
\cite{dataset_en_sentistrength} & en & SM &           Yes &    11759 &        28.0/34.0/38.0 &           26 &          147 \\
\cite{dataset_en_tweet_airlines} & en & SM &           Yes &    14427 &        63.0/21.2/15.8 &           17 &          104 \\
\cite{dataset_en_vader} & en & SM &            No &     4200 &        26.9/17.0/56.1 &           13 &           79 \\
\cite{dataset_es_paper_reviews} & es & M &            No &      163 &        33.7/33.7/32.5 &          135 &          835 \\
\cite{dataset_es_paper_reviews} & es & R &           Yes &      399 &        44.4/27.8/27.8 &          167 &         1020 \\
\cite{dataset_es_muchocine} & es & R &            No &     3871 &        32.9/32.3/34.9 &          511 &         3000 \\
\cite{dataset_fa_sentipers} & fa & R &           Yes &    13525 &        12.0/37.5/50.5 &           21 &          104 \\
\cite{dataset_he_hebrew_sentiment} & he & SM &           Yes &     8619 &         26.5/2.8/70.8 &           22 &          110 \\
\cite{dataset_hr_sentiment_news_document} & hr & N &           Yes &     2025 &        22.5/61.4/16.0 &          161 &         1021 \\
\cite{dataset_it_evalita2016} & it & SM &           Yes &     8926 &        36.7/41.7/21.6 &           14 &          101 \\
\cite{dataset_it_multiemotions} & it & SM &           Yes &     3139 &        24.4/14.9/60.6 &           17 &          106 \\
\cite{dataset_lv_ltec_sentiment} & lv & SM &           Yes &     5790 &        23.8/45.2/31.0 &           20 &          138 \\
\cite{dataset_pl_klej_allegro_reviews} & pl & R &            No &    10074 &        30.8/13.2/56.0 &           80 &          494 \\
\cite{dataset_pl_polemo} & pl & R &           Yes &    57038 &        42.4/26.8/30.8 &           30 &          175 \\
\cite{dataset_pl_opi_lil_2012} & pl & SM &           Yes &      645 &         50.7/47.3/2.0 &           33 &          230 \\
\footnotesize{\cite{dataset_pt_tweet_sent_br}} & pt & SM &           Yes &    10109 &        28.8/25.1/46.1 &           12 &           74 \\
\cite{dataset_ru_sentiment} & ru & SM &           Yes &    23226 &        16.8/54.6/28.6 &           12 &           79 \\
\cite{dataset_sl_sentinews} & sl & N &           Yes &    10417 &        32.0/52.0/16.0 &          309 &         2017 \\
\cite{dataset_sr_serb_movie_reviews} & sr & R &            No &     4724 &        17.8/43.7/38.5 &          498 &         3097 \\
\cite{dataset_sr_senticomments} & sr & R &           Yes &     3948 &        30.3/18.1/51.5 &           18 &          105 \\
\cite{dataset_th_wongnai_reviews} & th & R &            No &    46193 &         5.4/30.5/64.1 &           29 &          544 \\
\cite{dataset_th_wisesight_sentiment} & th & SM &           Yes &    26126 &        26.1/55.6/18.3 &            6 &           90 \\
\cite{dataset_ur_roman_urdu} & ur & M &           Yes &    19660 &        26.7/43.6/29.7 &           13 &           69 \\
\cite{dataset_ch_hotel_reviews} & zh & R &            No &   125725 &        28.6/21.9/49.5 &           51 &          128 \\
\end{tabular}
}
\end{table*}

\begin{table*}
\centering
\caption{List of all multilingual datasets used in experiments. Category (Cat.): R - Reviews, SM - Social Media, C - Chats, N - News, P - Poems, M - Mixed. HL - human labeled}
\label{tab:dataset_multilingual}
\begin{tabular}{c|c|c|c|r|r|r|r}
 Paper                     &   Cat. &   Lang  & HL &  Samples &  (NEG/NEU/POS) &  \#Words &  \#Char. \\
 \hline
\cite{dataset_dai_labor} & SM & de &           Yes &      953 &        10.0/75.1/14.9 &           12 &           80 \\
                                          &              & de &           Yes &     1781 &        16.9/63.3/19.8 &           13 &           81 \\
                                          &              & en &           Yes &     7073 &        17.4/60.0/22.6 &           14 &           78 \\
                                          &              & fr &           Yes &      685 &        23.4/53.4/23.2 &           14 &           82 \\
                                          &              & fr &           Yes &     1786 &        25.0/54.3/20.8 &           15 &           83 \\
                                          &              & pt &           Yes &      759 &        28.1/33.2/38.7 &           14 &           78 \\
                                          &              & pt &           Yes &     1769 &        30.7/33.9/35.4 &           14 &           78 \\
\cite{dataset_multilan_amazon} & R & de &            No &   209073 &        40.1/20.0/39.9 &           33 &          208 \\
                                          &              & en &            No &   209393 &        40.0/20.0/40.0 &           34 &          179 \\
                                          &              & es &            No &   208127 &        40.2/20.0/39.8 &           27 &          152 \\
                                          &              & fr &            No &   208160 &        40.2/20.1/39.7 &           28 &          160 \\
                                          &              & ja &            No &   209780 &        40.0/20.0/40.0 &            2 &          101 \\
                                          &              & zh &            No &   205977 &        39.8/20.1/40.1 &            1 &           50 \\
\cite{dataset_semeval_2017} & M & ar &           Yes &     9391 &        35.5/40.6/23.9 &           14 &          105 \\
                                          &              & en &           Yes &    65071 &        19.1/45.7/35.2 &           18 &          111 \\
\cite{dataset_semeval_2020} & SM & es &           Yes &    14920 &        16.8/33.1/50.0 &           16 &           86 \\
                                          &              & hi &           Yes &    16999 &        29.4/37.6/33.0 &           27 &          128 \\
\cite{dataset_twitter_sentiment} & SM & bg &           Yes &    62150 &        22.6/45.9/31.5 &           12 &           85 \\
                                          &              & bs &           Yes &    36183 &        33.4/30.5/36.1 &           12 &           75 \\
                                          &              & de &           Yes &    90534 &        19.7/52.8/27.4 &           12 &           94 \\
                                          &              & en &           Yes &    85784 &        26.8/44.1/29.1 &           12 &           77 \\
                                          &              & es &           Yes &   191412 &        11.8/37.9/50.3 &           14 &           92 \\
                                          &              & hr &           Yes &    75569 &        25.7/23.9/50.4 &           12 &           91 \\
                                          &              & hu &           Yes &    56682 &        15.9/31.0/53.1 &           11 &           83 \\
                                          &              & pl &           Yes &   168931 &        30.0/26.1/43.9 &           11 &           82 \\
                                          &              & pt &           Yes &   145197 &        37.2/35.0/27.8 &           10 &           61 \\
                                          &              & ru &           Yes &    87704 &        32.0/40.1/27.8 &           10 &           67 \\
                                          &              & sk &           Yes &    56623 &        25.6/22.5/51.9 &           13 &           97 \\
                                          &              & sl &           Yes &   103126 &        29.9/43.3/26.8 &           13 &           91 \\
                                          &              & sq &           Yes &    44284 &        15.7/33.1/51.1 &           13 &           90 \\
                                          &              & sr &           Yes &    67696 &        34.8/42.8/22.4 &           13 &           81 \\
                                          &              & sv &           Yes &    41346 &        40.3/31.2/28.5 &           14 &           94 \\
\end{tabular}
\end{table*}

\end{document}